# A Novel Fuzzy Bi-Clustering Algorithm with AFS for Identification of Co-Regulated Genes

Kaijie Xu

*Abstract*—The identification of co-regulated genes and their transcription-factor binding sites (TFBS) are the key steps toward understanding transcription regulation. In addition to effective laboratory assays, various bi-clustering algorithms for detection of the co-expressed genes have been developed. Bi-clustering methods are used to discover subgroups of genes with similar expression patterns under to-be-identified subsets of experimental conditions when applied to gene expression data. By building two fuzzy partition matrices of the gene expression data with the Axiomatic Fuzzy Set (AFS) theory, this paper proposes a novel fuzzy bi-clustering algorithm for identification of co-regulated genes. Specifically, the gene expression data is transformed into two fuzzy partition matrices via sub-preference relations theory of AFS at first. One of the matrices is considering the genes as the universe and the conditions as the concept, the other one is considering the genes as the concept and the conditions as the universe. The identification of the co-regulated genes (bi-clusters) is carried out on the two partition matrices at the same time. Then, a novel fuzzy-based similarity criterion is defined based on the partition matrixes, and a cyclic optimization algorithm is designed to discover the significant bi-clusters at expression level. The above procedures guarantee that the generated bi-clusters have more significant expression values than that of extracted by the traditional bi-clustering methods. Finally, the performance of the proposed method is evaluated with the performance of the three well-known bi-clustering algorithms on publicly available real microarray datasets. The experimental results are in agreement with the theoretical analysis and show that the proposed algorithm can effectively detect the co-regulated genes without any prior knowledge of the gene expression data.

## I. INTRODUCTION

Gene expression clustering allows an open-ended exploration of the data, without getting lost among the thousands of individual genes [1]. Traditional (global) clustering methods only analyze genes under all experimental conditions or only analyze conditions of all the genes. In practice, in numerous cellular processes, many genes are regularly co-expressed (co-regulated) [2] under some special conditions [3] but behave differently under different conditions. Consequently, mining local co-expressed valuable patterns becomes a vital objective in discovering genetic pathways that are not very clear when clustered globally [4]. This is the so-called bi-clustering problem. Bi-clustering extends the traditional clustering techniques by attempting to find (all) subgroups of genes with similar expression patterns under to-be-identified subsets of experimental conditions when applied to gene expression data.

Bi-clustering can discover co-regulated valuable patterns of gene from plenty of gene expression data, which are more helpful to define genes functioning together than traditional clustering approaches. The bi-clustering model measures coherence within the subset of genes and conditions. This model is effective in disclosing the involvement of genes or conditions in multipaths, some of which can only be uncovered under the dominance of more consistent ones [5]. The coherence score is usually defined through building a symmetric function of genes and conditions involved, and therefore the bi-clustering is a process of simultaneously clustering of genes and conditions. A so-called mean squared residue (MSR) defined by Cheng and Church [6]. is first introduced and applied to gene expression data transformed by a logarithm and augmented by the additive inverse. Furthermore, the MSR is also the most commonly used index in bi-clustering, and based on which many bi-clustering algorithms have been developed.

So far, various algorithms have been developed attempting to solve the bi-clustering problem. Popular bi-clustering algorithms, such as Cheng and Church (CC) algorithm [6], Flexible Overlapped Clusters (FLOC) [7], Plaid [8], order-preserving sub-matrix (OPSM) [9], Iterative Signature Algorithm (ISA) [10], conserved gene expression MOTIFs (xMOTIFs) [11], and BiMax [12] have drawn much attention in the literature. Emerging algorithms, such as Bayesian Bi-clustering [13], Maximum Similarity Bi-cluster algorithm (MSB) [14], and QUalitative BI-Clustering algorithm (QUBIC) [15] have not been extensively studied. In a word, as of now the research on the bi-clustering is still at its initial stage. The real power of this clustering strategy is yet to be fully realized due to the lack of effective and efficient algorithms for reliably solving the general bi-clustering problem.

Among the existing bi-clustering algorithms, the CC and the FLOC algorithms are considered as the most effective tools for processing gene expression data by now. The CC algorithm is the earliest and the most studied one, and the emerging algorithms are mostly based on the idea of the CC algorithm. The CC algorithm uses MSR of a bi-cluster as a similarity measure to greedily extract bi-clusters that satisfy a homogeneity constraint. It generates the row and column cluster randomly and then improves the bi-clusters to minimize the MSR value. Only one bi-cluster is identified each time and then replaced by random numbers before identifying the next cluster [16]. Based on this and with the aim of improving the generic CC algorithm, Yang et al. proposed another well-known method called FLOC [7], where an additional

K. Xu is with the School of Electronic Engineering, Xidian University, Xi'an 710071, China (e-mail: kjxu@xidian.edu.cn).

function is introduced to deal with the missing data and to discover the overlapping bi-clusters [17]. Subsequent studies suggest that the MSR is useful only for identifying certain classes of co-expressed genes, but not adequate to detect other transcriptionally co-regulated genes. Another well-known algorithm named QUBIC [15] which can solve the bi-clustering problem in a more general form, was proposed. The QUBIC algorithm can effectively and efficiently identify all statistically significant bi-clusters that cannot be identified by the other bi-clustering algorithms has turned out to be a more useful tool for identification of co-regulated genes.

Based on the Axiomatic Fuzzy Set (AFS) theory [18], this paper proposes a novel bi-clustering model for identification of co-regulated genes. AFS theory facilitates a way on how to transform data into fuzzy sets (membership functions) and implement their fuzzy logic operations, which provides a flexible and powerful tool for representing human knowledge and emulate human recognition process. In recent years, AFS theory has received increasing interest [18]. AFS theory takes uncertainty of randomness and imprecision of fuzziness as a unified and coherent process so that the membership functions are determined by the observed data. In AFS the fuzzy sets (membership functions) and their logic operations are impersonally and automatically determined a consistent algorithm according to the distributions of original data (AFS structures and AFS algebras) which is very different from the traditional fuzzy sets that the membership functions are often given by personal intuition and the logic operations are implemented by a kind of triangular norms (*t*-norm); the attributes of objects in it can be various data types or sub-preference relations, even human intuition descriptions; the distance function and objective function are not required, and any prior knowledge about the dataset is also not required. For a large dimensionality huge number of genes, it is impossible or difficult to define the membership functions just by personal intuition and define distance-based functions to implement fuzzy logic operations. Thus, this paper we design a bi-clustering algorithm to discover the co-regulated genes based on AFS theory.

From the design perspective, the sub-preference relations theory of AFS is used to build a fuzzy membership (**partition**) matrix only based on the distributions of original gene expression data, and it does not require any distance measures and the prior knowledge about the gene expression data. Specifically, in the proposed scheme, a reference gene is selected at first. Then, consider the genes as the universe and the conditions as the concept, a fuzzy partition matrix is built by the sub-preference relations theory [18]. Similarly, when considering the genes as the concept and the conditions as the universe, another fuzzy partition matrix can also be built. With the two partition matrices, we define a fuzzy-based similarity criterion to measure the similarity of the co-regulated genes under some special conditions. Subsequently, we design a cyclic optimization algorithm is to discover the bi-clusters (co-regulated genes). We believe that this is the first time that such a fuzzy-based similarity criterion is proposed and the first for solving bi-clustering problem. In addition, an approach based on the Fuzzy C-Means (FCM) [19-20] clustering is proposed to select a number of reference genes. Experimental studies completed on real-world gene expression data demonstrate that the proposed approach achieves better performance compared with that of the several well-known methods used for gene expression.

In brief, the major contribution of the paper is to propose a novel bi-clustering algorithm to discover the co-regulated genes. This algorithm does not require any prior knowledge about the gene expression data. To the best of our knowledge, the idea of the proposed approach has not been exposed in the previous studies.

This paper is organized as follows. The bi-clustering related concepts and novel similarity definitions based on the AFS theory and the principle of the proposed method are presented in Section II. Section III discusses the performance indexes. Section IV includes experimental setup and covers an analysis of completed experiments. Section V covers some conclusions.

## II. A Fuzzy Bi-clustering

### A. Problem Definitions

A commonly used way to visualize microarray data for gene expression analyses is to represent the data set as a matrix with rows representing the genes and columns representing the conditions (or the other way around) with each element of the matrix representing the expression value of a gene under a specific condition. Thus, identifying groups of genes in a microarray data set that share similar expression patterns under to-be-identified conditions is equivalent to finding submatrices with similar properties.

Let $A = [\ldots, a_{ij}, \ldots] \in R^{N \times M}$ be a microarray expression data matrix with a set of genes $G = [G_1, G_2, \ldots, G_N]^T$ and a set of conditions $O = [O_1, O_2, \ldots, O_M]$, where $a_{ij}$ represents an expression value of a gene $G_i$ under a condition $O_j$. A bi-cluster is basically a sub-matrix $A_{IJ}$ that exhibits some similar tendency, which can be expressed by $A(I, J)$, where $I \subset N$ and $J \subset M$ are subsets of genes and conditions, respectively. Let $g^* \in G$ be a reference gene, our goal is to find a subset of genes (co-regulated genes, bi-cluster) that are related to $g^*$. When the reference gene is not known, we can enumerate all genes in the matrix or randomly select several genes as the reference genes subsets. Similar ideas are also used in [21].

### B. Similarity Definitions

Consider the aforementioned gene expression data matrix $A$. First, we use the AFS theory to build two fuzzy partition matrices [22] which are used to define the similarity matrices in this paper. One of the matrices is considering the genes as the universe and the conditions as the concept, the other one is considering the genes as the concept and the conditions as the universe. Assume that the two fuzzy partition matrices are $U_G = [\ldots, \mu_{ij}, \ldots]$ and $U_C = [\ldots, u_{ij}, \ldots]$. The calculation of the membership degree based on the AFS is determined as follows [18, 23]:

$$\mu(x) = \sup_{i \in I} \left\{ \frac{\mathcal{M}[A_i(x)]}{\mathcal{M}(x)} \right\} \quad (1)$$

where $x \in X$, $X$ is the universe of discourse, $\mathcal{M}$ is a finite and positive measure over $\sigma$-algebra, $I$ is a non-empty indexing set [23].

For a gene expression data matrix $A_{IJ}$ and a given reference gene $g^* \in G$, define $U_{Gg^*} = U_G - \mu_{g^*} = [K, \delta_{ij}, L]$, $\delta_{ij} = |\mu_{ij} - \mu_{g^*j}|$. Obviously, $\delta_{ij}$ can characterize the similarity between the $i$th gene and the reference gene under the $j$th condition, and the smaller the $\delta_{ij}$ is, the larger the similarity is, and vice versa. Furthermore, $U_C$ is used as another similarity to jointly discover the local co-regulated genes, and this is the so-called the proposed bi-similarity.

Let $U_{Gg^*}(N,M)$ be an $N \times M$ gene similarity matrix and $U_{Gg^*}(I,J)$ be a bi-cluster (sub-matrix) of $U_{Gg^*}(N,M)$. For column $j \in J$, we define the dissimilarity score of the $j$-column in $U_{Gg^*}(I,J)$ is the Range of the column, i.e.,

$$\mu(I,j) = \sum_{i \in I}\left[\max(\mu_{I,j}) - \min(\mu_{I,j})\right] \quad (2)$$

The dissimilarity score of $U_{Gg^*}(I,J)$ is

$$\mu(I,J) = \frac{1}{J}\sum_{(i \in I), j=1}^{J}\left[\max(\mu_{I,j}) - \min(\mu_{I,j})\right] \quad (3)$$

Let $U_C(N,M)$ be an $N \times M$ condition similarity matrix and $U_C(I,J)$ be a bi-cluster (sub-matrix) of $U_C(N,M)$. For row $i \in I$, we define the dissimilarity score of the $i$-row in $U_C(I,J)$ is the Range of the row, i.e.,

$$u(i,J) = \sum_{j \in J}\left[\max(u_{i,J}) - \min(u_{i,J})\right] \quad (4)$$

The dissimilarity score of $U_C(I,J)$ is

$$u(I,J) = \frac{1}{I}\sum_{(j \in J), i=1}^{I}\left[\max(u_{i,J}) - \min(u_{i,J})\right] \quad (5)$$

Consider a bi-cluster $A(I,J)$. If the dissimilarity score $U_{Gg^*}(I,J)$ and is low, and the genes under all the conditions in $A(I,J)$ have similar expression values. However, many genes are regularly co-expressed under some special conditions, in other words, mining local co-expressed valuable patterns is more meaningful than that of clustering globally. Thus, we use $U_{Gg^*}$ and $U_C$ to jointly discover the local co-expressed valuable patterns. For a bi-cluster $A(I,J)$, if both $U_{Gg^*}\bar{U}_G(I,J)$ and $U_C(N,M)$ are low, then the genes in $A(I,J)$ under the $J$ conditions are co-expressed.

Based on the above analysis, we will report the proposed algorithm of discovering the bi-clusters.

### C. Bi-clusters Discovery

The algorithm is an essentially greedy algorithm, and it starts with the whole gene expression data matrix $A(N,M)$ as an initial bi-cluster. In the discovering of bi-clusters, we define a novel bi-similarity criterion based on the similarity matrices above. With the use of the bi-similarity criterion, the co-regulation of the genes in the same bi-clusters become enhanced.

Let $\mu(N,M)$ and $u(N,M)$ be the dissimilarity scores of $U_{Gg^*}(N,M)$ and $U_C(N,M)$, respectively. our goal is to discover a bi-cluster with small dissimilarity scores, such as $\mu(N,M)/\alpha$ and $u(N,M)/\beta$, where $\alpha$ and $\beta$ are the scale factors of the column and row of the bi-cluster. Thus, we can call the bi-cluster an (αβ)-bi-cluster. An excellent bi-cluster is generated by deleting and adding rows and columns with some particular rules. The sketch is as follows:

**Algorithm 1 (Node Deletion)**

**Input:** $U_G$ and $U_C$, the two fuzzy partition matrices of the gene expression data matrix, and $\alpha, \beta$, the two scale factors of the column and row for the bi-clusters to be found.

**Output:** $A(I,J)$, an (αβ)-bi-cluster that is a sub-matrix of $A(I,J)$ with row set $I$ and column set $J$ with the dissimilarity scores no larger than $\mu(N,M)/\alpha$ and $u(N,M)/\beta$, respectively.

**Initialization:** $I$ and $J$ are initialized to the gene and condition sets in the gene expression data, and $A_{IJ} = A$; a reference gene $g^*$ is given by the user; the maximum acceptable dissimilarity scores of the column and row: $\mu(N,M)/\alpha$, $u(N,M)/\beta$.

**Iteration:**

1. Calculate $\mu(I,j)$ for all $j \in J$, $u(i,J)$ for all $i \in I$, and $\mu(I,J)$, $u(I,J)$. If $\mu(I,J) \leq \mu(N,M)/\alpha$ and $u(I,J) \leq u(N,M)/\beta$, return $A_{IJ}$.

2. Find column $j \in J$ with largest

$$\mu(I,j) = \sum_{i \in I}\left[\max(\mu_{I,j}) - \min(\mu_{I,j})\right] \quad (6)$$

and row $i \in I$ with largest

$$u(i,J) = \sum_{j \in J}\left[\max(u_{i,J}) - \min(u_{i,J})\right] \quad (7)$$

remove the column if

$$\frac{\alpha\mu(I,j)}{\mu(N,M)} > \frac{\beta u(i,J)}{u(N,M)} \quad (8)$$

else remove the row by updating either $I$ or $J$.

Clearly, after node deletion, both the row and column dissimilarity scores of the sub-matrix will be reduced. However, the resulting (αβ)-bi-cluster may not be maximal, in the sense that some rows and columns may be added without increasing the dissimilarity scores. Thus, we design another algorithm to refine the bi-clusters.

**Algorithm 2 (Node Addition)**

**Input:** $A_{IJ}$, a sub-matrix of real numbers, $I$ and $J$ signifying an (αβ)-bi-cluster.

**Output:** $A(I,J)$, $I'$ and $J'$ such that $I' \subset I$ and $J' \subset J$ with the property that

$$\mu(I',J') \leq \mu(I,J) \ \& \ u(I',J') \leq u(I,J) \quad (9)$$

**Iteration:**

1. Compute $\mu(I,j)$ for all $j \notin J$, and recompute $\mu(I,J)$ and $u(i,J)$, and add the columns $j \notin J$ if $\mu(I,J) \leq \mu(N,M)/\alpha$.

$$\mu(I,J) \leq \frac{\mu(N,M)}{\alpha} \ \& \ u(I,J) \leq \frac{u(N,M)}{\beta} \quad (10)$$

2. Compute $u(i,J)$ for all $i \notin I$, and recompute the $u(I,J)$ and $\mu(I,J)$, and add the rows $i \notin I$ if

$$\mu(I,J) \le \frac{\mu(N,M)}{\alpha} \ \& \ u(I,J) \le \frac{u(N,M)}{\beta} \qquad (11)$$

3. If nothing is added in the iterate, return the final $I$ and $J$ as $I'$ and $J'$.

Obviously, after the execution of the node addition algorithm, neither the row dissimilarity score nor column dissimilarity score will increase. Sometimes an addition may decrease the score more than any deletion.

### D. Selection of the Reference Genes

In the algorithm proposed above, the reference genes we are interested in are known in advance. When the reference genes are unknown, we should select a number of genes as the reference genes. In some cases, the reference genes selected is closely related to the quality of the bi-clusters. Furthermore, we usually wish the size (number of the co-regulated genes and conditions) of the bi-clusters to be as large as possible (discover more co-regulated genes under more conditions). In other words, if a gene has more similar genes under more conditions, then it is more suitable to be a reference. Based on this, we propose a method to select the reference genes.

Firstly, we calculate the fuzzy similarity matrices of all genes under all each ($j$th) condition, and we obtain $M$ similarity matrices, say,

$$S_j = \left[ s_1^{(j)}, s_2^{(j)}, \ldots, s_i^{(j)} \ldots, s_N^{(j)} \right]$$
$$s_i^{(j)} = \left[ s_1^{(ji)}, s_2^{(ji)}, \ldots, s_k^{(ji)}, \ldots, s_N^{(ji)} \right]^T \qquad (12)$$
$$i = 1, 2, \ldots, N; \ j = 1, 2, \ldots, M; \ k = 1, 2, \ldots, N$$

where $T$ stands for the transpose operation. Let $V_i^{(j)} = [\max(s_i^{(j)}), \min(s_i^{(j)})]^T$ be the prototypes of $s_i^{(j)}$, based on FCM clustering we transform $s_i^{(j)}$ into a membership matrix as follows:

$$\boldsymbol{\Phi}_i^{(j)} = \left[ \varphi_{k1}^{(ji)} \ \varphi_{k2}^{(ji)} \right]^T \in R^{2 \times N}$$
$$\varphi_{kc}^{(ji)} = \frac{\left\| s_k^{(ji)} - v_c^{(ji)} \right\|^{\frac{-2}{m-1}}}{\sum_{h=1}^{2} (\frac{1}{\left\| s_h^{(ji)} - v_c^{(ji)} \right\|})^{\frac{2}{m-1}}} \qquad (13)$$
$$c = 1, 2; \ k = 1, 2, \ldots, N$$

where $v_c^{(ji)}$ is the $c$-th element (cluster prototype) of $V_i^{(j)}$; $m$ is a fuzziness exponent (fuzziness coefficient); $\|\cdot\|$ stands for the Euclidean distance. $\varphi_{kc}^{(ji)} \in [0, 1]$ is the degree of membership of individual $s_k^{(ji)}$ belonging to the cluster $c$, and satisfies the following condition

$$\sum_{c=1}^{2} \varphi_{kc}^{(ji)} = 1 \ \text{for} \ k = 1, 2, \ldots, N \qquad (14)$$

Then, we construct a function to compute the mean of the large fuzzy similarity of the $i$th gene under all the $M$ conditions.

$$\varpi_i = \frac{\sum_{j=1}^{M} \sum_{k=1}^{N} s_k^{(ji)} sign\left[ \varphi_{k1}^{(ji)} - 0.5 \right]}{\delta \left\{ \sum_{k=1}^{N} sign\left[ \varphi_{k1}^{(ji)} - 0.5 \right] \right\} + M \sum_{k=1}^{N} sign\left[ \varphi_{k1}^{(ji)} - 0.5 \right]} \qquad (15)$$

where $\delta$ is a unit pulse response function. Generally, we consider that the genes with large $\varpi$ values are more suitable to be references, since they have more similar genes under more conditions as previously described.

## III. PERFORMANCE INDEXES

In order to evaluate the performance of the proposed algorithm, two commonly used performance indexes are briefly discussed.

### A. Variance Index

Given a gene expression data matrix $A(N,M)$, with set of rows (genes) $N$ and set of columns (conditions) $M$, a bi-cluster is a sub-matrix $A(I,J), I \subset N, J \subset M$ of $A(N,M)$. $a_{ij}$ is the value in the data matrix $A$ corresponding to row $i$ and column $j$. We denote by $a_{iJ}$ the mean of the $i$th row in the bi-cluster, $a_{Ij}$ the mean of the $j$th column in the bi-cluster and $a_{IJ}$ the mean of all elements in the bi-cluster. These values are defined by:

$$a_{iJ} = \frac{1}{|J|} \sum_{j \in J} a_{ij} \qquad (16)$$

$$a_{Ij} = \frac{1}{|I|} \sum_{i \in I} a_{ij} \qquad (17)$$

$$a_{IJ} = \frac{1}{|I||J|} \sum_{i \in I, j \in J} a_{ij} = \frac{1}{|I|} \sum_{i \in I} a_{iJ} = \frac{1}{|J|} \sum_{j \in J} a_{Ij} \qquad (18)$$

The variance [24] is used to evaluate the quality of each bi-cluster $A(I,J)$:

$$VAR(I,J) = \sum_{i \in I, j \in J} \left( a_{ij} - a_{IJ} \right)^2 \qquad (19)$$

the lower the value returned, the better the quality of the bi-cluster will be; a perfect bi-cluster is a sub-matrix with variance equal to zero.

### B. Mean Fluctuation Degree Index

Mean Fluctuation Degree (MFD) is used to evaluate the changing trends of the genes under each condition transition. the MFD of a bi-cluster is defined as

$$MFD(I,J) = \sqrt{\frac{1}{|I||J|} \sum_{i \in I, j \in J} \left( \Theta_{ij} - \frac{1}{|I|} \sqrt{\Theta_{ij}} \right)^2} \qquad (20)$$

where

$$\Theta_{ij} \in \boldsymbol{\Theta}$$
$$\boldsymbol{\Theta} = 180 \arctan\left( \Delta^\dagger \boldsymbol{\Xi} \right) / \pi \qquad (21)$$

$$\Delta = \frac{diag\left\{ \max(a_{1j}) - \min(a_{1j}), \ldots, \max(a_{ij}) - \min(a_{ij}), \ldots \right\}}{M-1} \qquad (22)$$
$$i = 1, 2, \ldots, N; \ j = 1, 2, \ldots, M$$

$$\boldsymbol{\Xi} = [\ldots, a_{ij} - a_{i(j-1)}, \ldots] \in R^{N \times (M-1)} \qquad (23)$$
$$i = 1, 2, \ldots, N; \ j = 2, \ldots, M$$

where subscript † denotes the Moore-Penrose Inverse of the matrix [25]. Obviously, for a bi-cluster if the genes (rows) have similar changing trends under each condition transition, its MFD will be relatively smaller. Furthermore, if all genes (rows) in the bi-cluster have the completely similar (or same) changing trends under each condition transition, its MFD will be zero.

In particular, for a single-row (or a single-column) "bi-cluster", its VAR and MFD indexes are also zero, however, such "bi-cluster" is meaningless. To fairly compare the performance of the algorithms, we will drop the resulting bi-clusters with only one row and one column.

IV. EXPERIMENTAL STUDIES

In the following experiments, we compare the performance of the proposed fuzzy bi-clustering (FBC) method with CC, FLOC and QUBIC methods, which are the two well-known bi-clustering methods commonly used for gene expression. In the experiments, two well-known publicly available real microarray datasets named Yeast (http://arep.med.harvard.edu/biclustering/yeast.matrix) and Gordon 2002 [26]. which are the most commonly used datasets in bi-clustering. Both the variance [24] and the mean fluctuation degree are taken as the evaluation indexes which are briefly discussed as follows.

A. Experiments

The methods are used to find 100 bi-clusters. A concise description the values of the parameters used in the experiments are given in Table I. The methods are repeated 10 times; the means and the standard deviations of the experimental results are presented. The experimental results are plotted in Figs. 1 to 2. It is clear that the proposed algorithm is effective in discovering the quality of bi-clusters by grouping together genes that have trends with more similar fluctuation degrees. Compared with the CC, FLOC and QUBIC clusters methods, the proposed methods exhibit visible advantages. The improvement is 25% (on average) and varies in-between a minimal improvement of 15% and 35% in the case of the most visible improvement.

TABLE I. Datasets and parameters used in the experiments

| Datasets | | Yeast | Gordon-2002 |
|---|---|---|---|
| Number of | Genes | 2,884 | 1,626 |
| | Conditions | 17 | 181 |
| Threshold of MSR | | 300 | 3,000 |
| α | | 5.0 | 5.5 |
| β | | 1.8 | 3.0 |

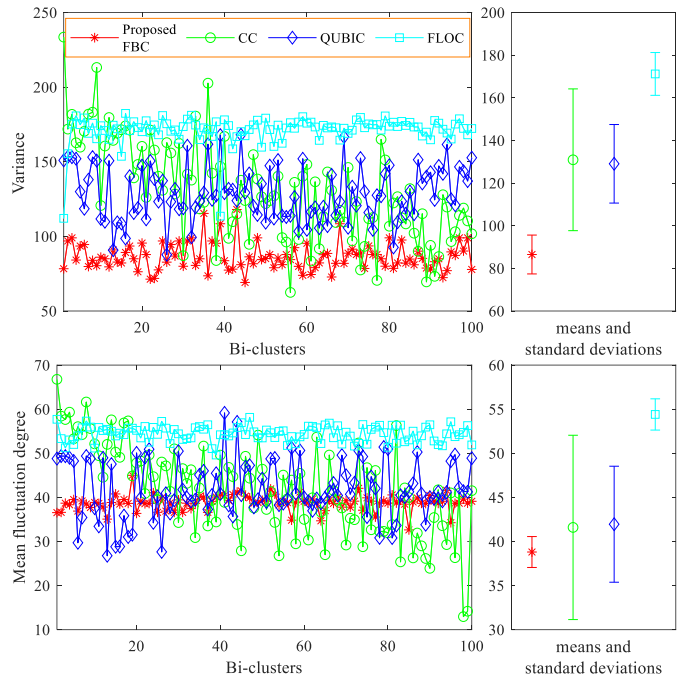

Fig. 1. Comparison of the performance of methods on the Yeast dataset

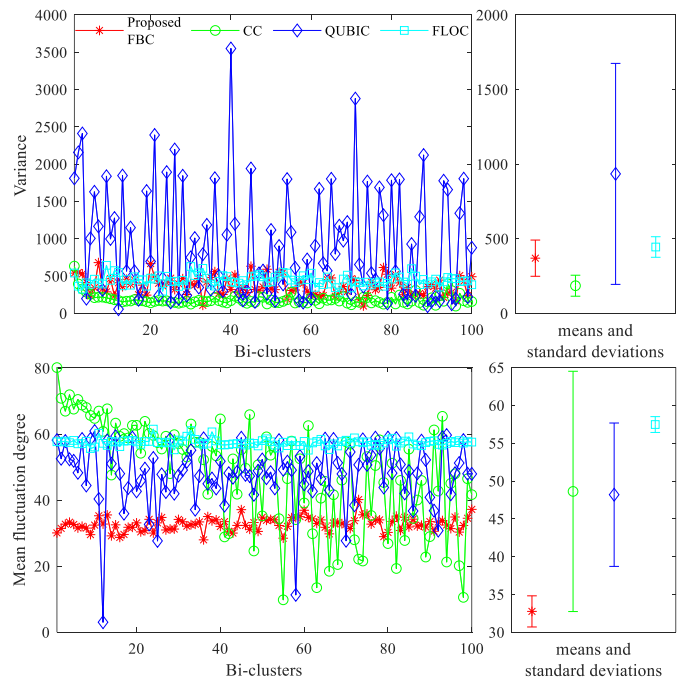

Fig. 2. Comparison of the performance of methods on the Gordon 2002 dataset

V. CONCLUSIONS

In this research, we designed a bi-clustering algorithm for identification of co-regulated genes via AFS theory. During the design process, the sub-preference relations theory of AFS is introduced to construct two fuzzy membership matrices to define a fuzzy-based similarity criterion. With the similarity criterion, a cyclic optimization algorithm is designed to discover the bi-clusters (co-regulated genes). We conducted theoretical analysis and offered a comprehensive suite of experiments. Both the theoretical and experimental results are presented to verify the validity of the proposed method. Experimental results show that the proposed method outperforms the existing algorithms in finding the bi-clusters, and has demonstrated its outstanding performance and great potential for development of gene expression. To the best of our knowledge, this research scheme is first proposed, which steadily improves the performance of the bi-clustering. At the current stage, we have completed a theoretical analysis and offered a comprehensive suite of experiments. Some practical experiments would be an interesting direction to explore in the future studies.